\documentclass[letterpaper,10pt,conference]{IEEEtran}

\usepackage{copyright}

\usepackage{graphicx}
\usepackage{tabularx}
\usepackage[font=normalsize]{caption}
\usepackage{subcaption}
\usepackage{xcolor,colortbl}
\usepackage{amsmath}
\usepackage{amssymb}
\usepackage{algorithm}
\usepackage[noend]{algpseudocode}
\usepackage{bbm}
\usepackage{hyperref}
\usepackage{cleveref}
\usepackage[inline]{enumitem}
\usepackage{siunitx}

\usepackage{cleveref}
\crefname{table}{Tab.}{Tabs.}
\crefname{figure}{Fig.}{Figs.}
\crefname{section}{Sec.}{Secs.}
\crefname{equation}{Eq.}{Eqs.}

\DeclareMathOperator*{\argmin}{arg\,min}

\title{Explainable Action Prediction through Self-Supervision on Scene Graphs}

\author{Pawit Kochakarn, Daniele De Martini, Daniel Omeiza, Lars Kunze\\
Oxford Robotics Institute, University of Oxford, UK\\
\texttt{\{pkochakarn, daniel\}@oxfordrobotics.institute} \quad \texttt{\{daniele, lars\}@robots.ox.ac.uk}}

\begin{document}
\maketitle

\copyrightnotice

\begin{abstract}
This work explores scene graphs as a distilled representation of high-level information for autonomous driving, applied to future driver-action prediction.
Given the scarcity and strong imbalance of data samples, we propose a self-supervision pipeline to infer representative and well-separated embeddings.
Key aspects are interpretability and explainability; as such, we embed in our architecture attention mechanisms that can create spatial and temporal heatmaps on the scene graphs.
We evaluate our system on the ROAD dataset against a fully-supervised approach, showing the superiority of our training regime.
\end{abstract}

\begin{figure*}[h!]
\centering
\includegraphics[width=0.85\textwidth]{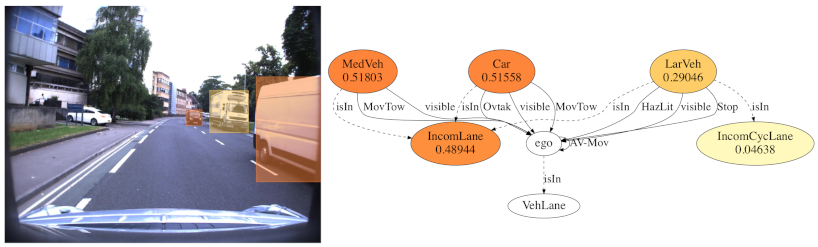}
\caption{Spatial (node) attention scores from a correctly predicted \emph{Move} scene.
Notably, the model paid less attention to a parked truck (\emph{LarVeh}) with its hazard lights on in the incoming cycle lane compared to the dynamic agents in the oncoming lane.\label{fig:node_attn_viz}}
\end{figure*}

\section{Introduction} \label{intro}
Autonomous Driving (AD) has been a popular and fruitful research field over the past two decades; still, there is a huge performance gap between a human driver and an autonomous vehicle (AV) when attaining a higher-level understanding of complex driving scenes, especially within urban environments.
This lack of high-level understanding can lead to the vehicle not perceiving or planning for hidden risks and potentially to accidents or user disengagements.
Furthermore, explainability is a crucial feature in modern AV systems: to gain trust and confidence from users, AVs should be able to explain what they have \textit{seen}, \textit{done} and \textit{might do} \cite{Omeiza_2021}.

Research has suggested that humans rely on cognitive mechanisms for representing structure and reasoning about inter-object relations when performing complex tasks \cite{relational_bias}.
While modern AV perception proficiently detects objects and road geometry and estimates vehicle trajectories, it does not explicitly capture inter-object relations that are key to understanding scenes.
Many researchers have suggested using a knowledge-graph variant, \emph{scene-graphs}, to model spatio-temporal relationships between agents and the road state.
Scene graphs act as structured representations of scenes: objects and their attributes are encoded as nodes connected by edges representing pairwise relationships, giving a very compressed but informative scene abstraction.


Scene graphs have been beneficial in modelling driving scenes for tasks such as vehicle behaviour classification \cite{vehicle_cls}, collision prediction \cite{collision_pred} and risk assessment \cite{risk_assess}.
However, these were formulated as supervised-learning problems, where driving clips are manually labelled to learn the specific downstream task.
On the other hand, approaches that rely on unsupervised or self-supervised methods to learn robust representations of scene graphs are limited.
The reason for attempting this is to try to address the shortcomings that come with a reliance on labels for downstream tasks.
Manual annotation, especially for large-scale datasets and high-level information, is expensive and purely supervised learning methods suffer from poor generalisation due to over-fitting -- mainly when training data is scarce -- and noisy labels~\cite{graph-ssl}.

The main contributions of this paper include:
\begin{enumerate*}
  \item Design an encoder network that can learn embeddings from a sequence of scene graphs in a self-supervised manner;
  \item Embed attention mechanisms in the encoder to foster explainability and help interpret model decisions;
  \item Evaluate the learnt embeddings and attention masks on the task of future driver-action prediction on real-world data.
\end{enumerate*}


\section{Related Work}

\paragraph{Scene Graphs for Autonomous Driving}
There have been several works on graph-based driving scene understanding.
\cite{3d-aware-scenegraph} proposed using multi-relational graph convolutional networks to model ego-centric spatio-temporal interactions through an \emph{Ego-Thing} and \emph{Ego-Stuff} graph to encode ego-vehicle interactions with \emph{thing}s (e.g. cars and pedestrians) and \emph{stuff} (e.g. lane markings and traffic lights).
\cite{dynamic_scene_understanding} similarly model the temporal evolution of spatial relations and thus predict the vehicle behaviour of each dynamic agent in the scene.
Finally, \cite{malawade2022roadscene2vec} proposed a general-purpose scene-graph learning library called \emph{roadscene2vec} for evaluating different Graph Neural Networks (GNNs) on downstream tasks such as risk assessment and collision prediction.
%
Our proposed method aligns most with \emph{roadscene2vec} \cite{malawade2022roadscene2vec} in that each spatio-temporal scene graph consists of a sequence of static scene graphs over a finite time horizon.

\paragraph{GNNs for Scene Graph Learning}
Recently, GNNs have grown to be very successful at prediction, classification and recommendation tasks for an increasing number of applications, from drug discovery to social networks \cite{gnn-survey}.
In scene understanding,
\cite{group-activity-recog} apply them to group activity recognition and \cite{scene-graph-gen} to dynamic scene-graph generation.
%
For example \cite{action-recog} perform action recognition from raw video input using scene graph sequences.
In our work, a similar method is used in encoding a single \emph{spatio-temporal embedding} from a sequence of scene graphs.
However, the way the embedding is learnt and the respective downstream task differ.

\paragraph{Graph Self-Supervised Learning}
A core motivation behind our work is to learn representations of driving scenes without the need for supervision.
Self-supervised learning (SSL) has attracted significant interest lately to reduce dependence on manual labels, but learning on pseudo-labels generated directly from the data, leading to more generalised representations and bolstering downstream tasks~\cite{graph-ssl}.
In the context of graph SSL, Graph Contrastive Learning (GCL) is gaining popularity: GraphCL \cite{graph-cl}, GRACE \cite{grace} and InfoGraph \cite{infograph} rely on standard data augmentation methods, contrasting modes and objective functions to perform node or graph classification.
We take inspiration from  them and SSL frameworks used for images (e.g. SimCLR \cite{simclr}) to perform representation learning on the constructed driving scene graphs.

\paragraph{Explainability for Scene Understanding}
Providing explainability in AV systems is beneficial for building public trust and, more importantly, streamlining the process of verifying and debugging predictions~\cite{Omeiza_2021}.
%
\cite{explain-gcn} proposed three methods inspired by the explainability work done on CNNs: gradient-based heat maps, Class Activation Mapping (CAM), and Excitation Backpropagation (EB), all aiming to generate heatmaps over the input data to highlight specific areas relevant to the model's decisions.
They showed that a variant of the CAM method (Grad-CAM) is most suitable for explaining general moderate-sized graphs.
Our work, instead, will use the concept of spatio-temporal attention from \cite{malawade2022roadscene2vec} to generate explanations for the primary use case of verifying and debugging predictions.
As, differently than Grad-CAM, it doesn't require post-processing of layer activations and gradients while also providing temporal explanations.


\section{Methodlogy}

\paragraph{Scene Graphs}
Let's consider a spatio-temporal scene graph as an attributed dynamic graph with node and edge features.
Let \(\mathcal{G}^{(t)} = (X^{(t)}_{node}, A^{(t)}, X^{(t)}_{edge})\) be a dynamic graph indexed at time \(t \in \mathbb{R}^{+}\), where the node feature matrix \(X^{(t)}_{node} \in \mathbb{R}^{N \times d_{node}}\), adjacency matrix \(A^{(t)} \in \mathbb{R}^{N \times N}\) and edge feature matrix \(X^{(t)}_{edge} \in \mathbb{R}^{N \times d_{edge}}\) all evolve with respect to time.
Here, $N$, $d_{node}$ and $d_{edge}$ are the number of nodes in the graph and the feature dimensions for nodes and edges.
As a finite sequence of n-graphs is used, it is more convenient to use \(\mathcal{G}_{1\dots n} = \{\mathcal{G}^{(1)}, \dots ,\mathcal{G}^{(n)}\}\) to represent each spatio-temporal scene graph.

\subsection{Encoder Architecture} \label{encoder}
\begin{figure}
\centering
\includegraphics[width=0.16\textwidth]{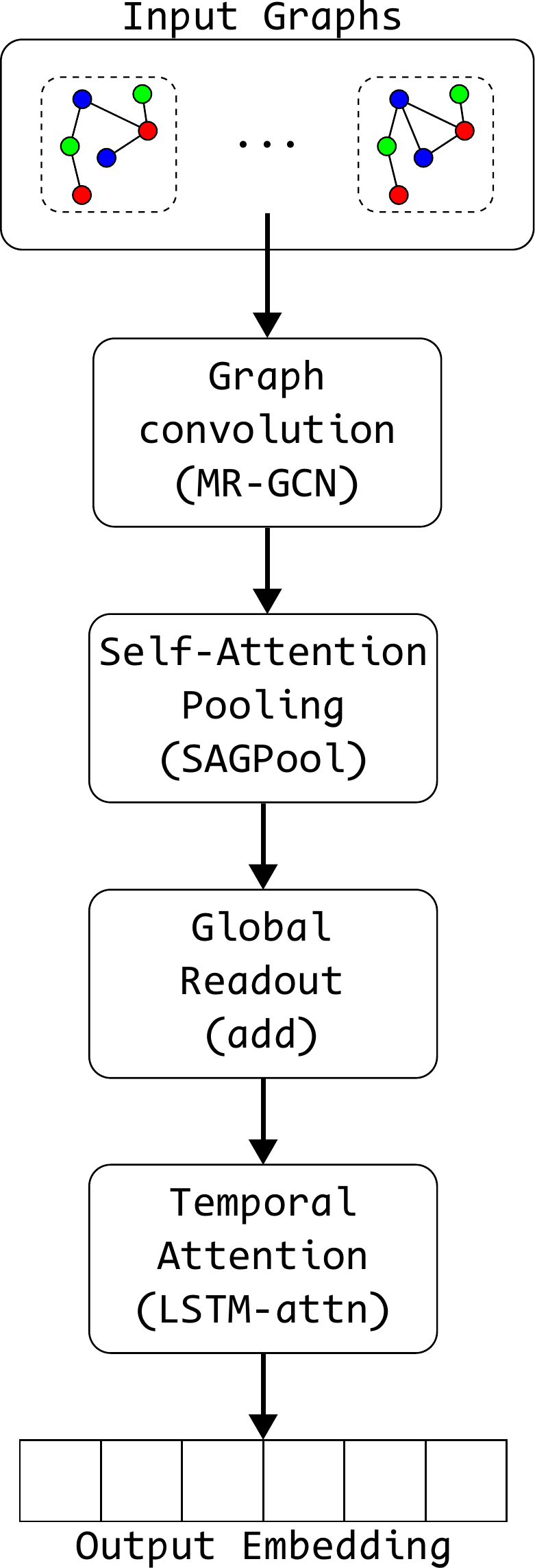}
\caption{Graph Encoder \(f_{\theta}\) architecture\label{fig:encoder_arch}}
\end{figure}

We design the graph encoder (\cref{fig:encoder_arch}) to allow for spatial and temporal modelling of scene graphs.
The key spatial components include (i) graph convolution layers, (ii) graph attention pooling layers, and (iii) graph readout layers.

\textbf{Multi-Relational Graph Convolution (MR-GCN):} A sequenc $\Tilde{\mathcal{G}}_{1...n}$ is fed to a multilayer MR-GCN layers to obtain a \emph{K}-hop spatial representation for each node in terms of its neighbours.
We base our approach on \emph{graph isomorphism networks}~\cite{gin}.

\textbf{Self-Attention Graph Pooling (SAGPool):} We ustilise SAGPool~\cite{sagpool} -- a form of hierarchical pooling which uses graph features, topology and self-attention -- on the produced node embeddings to extract only those most beneficial for the learning task.

\textbf{Global Readout:} We perform a readout operation (\texttt{add} in our case) over each set of node embeddings to output a full graph embedding of fixed dimensions then passed through a linear layer.
Its output is a sequence of embeddings for each scene graph in the input.
%

\textbf{Temporal Attention (LSTM-attn):} The final component in the encoder consists of using an LSTM with attention to convert the sequence of graph embeddings into a single spatio-temporal embedding, often called a \emph{context vector}.
We use an attention mechanism to allow the network to ``focus'' on embeddings in the sequence that matter most to the overall scene context: taking inspiration from~\cite{seq2seq}, we compute weights for each embedding using a feed-forward layer on the LSTM output and final hidden state.
The final spatio-temporal embedding (\emph{context} vector) is thus an attention-weighted combination of hidden states over the whole sequence; it will be used in further downstream tasks such as classification or prediction.

\subsection{Graph Contrastive Learning\label{sec:contr_learning}}

GCL is based on maximising mutual information between two augmented instances of the same object (e.g. node, subgraph or graph).
The framework can be formalised as an encoder-decoder network, where an encoder \(f_{\theta}\) learns a low-dimensional representation (or embedding) from each graph sequence \(\Tilde{\mathcal{G}}_{1...n}\).
A pretext decoder \(p_{\phi}\) then acts as a discriminator for estimating agreement between representations. The learning objective can be formulated as:
\begin{equation}
\label{eqn:gcl_formulation}
\theta^{*}, \phi^{*} = \argmin_{\theta, \phi} \mathcal{L}_{con}\bigg(p_{\phi}\Big(f_{\theta}(\Tilde{\mathcal{G}}_{1...n}^{(1)}), f_{\theta}(\Tilde{\mathcal{G}}_{1...n}^{(2)})\Big)\bigg),
\end{equation} where \(\Tilde{\mathcal{G}}_{1...n}^{(1)}\) and \(\Tilde{\mathcal{G}}_{1...n}^{(2)}\) are two different augmented instances of \(\mathcal{G}_{1...n}\), and \(\mathcal{L}_{con}\) denotes a contrastive loss. After training the encoder \(f_{\theta}\) to obtain optimal parameters \(\theta^{*}\), it can then bootstrap the training process of a downstream task in a supervised setting:
\begin{equation}
 \label{eqn:gcl_sup_formulation}
 \theta^{**}, \psi^{*} = \argmin_{\theta^{*}, \psi} \mathcal{L}_{sup}\bigg(q_{\psi}\Big( f_{\theta^{*}}(\mathcal{G}_{1...n})\Big), y \bigg),
\end{equation} where \(q_{\psi}\) is a downstream decoder,  \(y\) the downstream task labels, and \(\mathcal{L}_{sup}\) a supervised loss.

\paragraph{Data Augmentation}
Contrastive learning relies heavily on well-crafted augmentation.
Taking inspiration from GCL literature~\cite{graph-cl}, two stochastic augmentation methods were devised that fit within the driving context, to chain together and apply on a per-scene-graph basis instead of whole sequences.
The first method randomly drops \texttt{visible} agent nodes from each scene graph in $\mathcal{G}_{1...n}$, as we argue that agents located at a safe distance don't contribute as much to a scene's representation as closer agents.
This method fosters the model robustness to noisy driving scenes (e.g. busy junctions) when performing a downstream task.
The second augmentation randomly permutes  edges in the graph to either one proximity class above or below to bolster robustness to inaccuracies in depth estimation and help the model better generalise to ambiguities in semantic depth estimation.

While other graph augmentation techniques exist in the literature \cite{graph-ssl} (e.g. node feature masking and shuffling, graph diffusion or random-walk-based subgraph sampling), these were found to corrupt the scene graph too much where learning results became sub-optimal.

\paragraph{Loss}
After obtaining the two spatio-temporal embeddings \(g_{1}, g_{2}\), these are projected into a contrastive embedding space using a pretext decoder \(p_\phi\), $z = p_\phi(g)$.
In this work, \(p_\phi\) is a Multi-Layer Perceptron (MLP) with one hidden layer and normalisation, in line with GraphCL \cite{graph-cl} and SimCLR \cite{simclr}.
We chose to use InfoNCE loss, among others such as the Jensen-Shannon Estimator (JSE)~\cite{jse}, as these foster node-level tasks rather than graph-level tasks.

\section{Experiments}

\begin{figure}
\begin{subfigure}{0.44\textwidth}
\centering
    \includegraphics[width=0.95\textwidth]{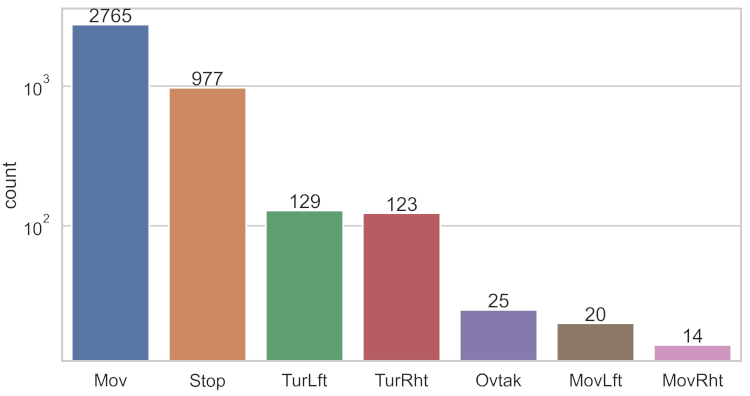}
    \caption{\label{fig:class_count}}
\end{subfigure}

\begin{subtable}{0.43\textwidth}
    \centering
    \renewcommand{\arraystretch}{1.3}
    \small
    \begin{tabular}{c|c|c}
       RID & Depth [m] &  Proximity \\ \hline
       0 - 0.15   &  $>$ 10 & \texttt{visible}   \\ 
       0.16 - 0.30 & 5 - 10  &   \texttt{near} \\ 
       0.31 - 1.0 & 0 - 5  &  \texttt{near\_collision}  \\
    \end{tabular}
    \caption{\label{table: depth labels}}
\end{subtable}
\caption{Count plot with log-scale to show class imbalance within the dataset.}
\vspace{-15pt}
\end{figure}

\paragraph{Downstream Learning Task}
We apply the learnt embeddings to the downstream task of \emph{driver action prediction}, defined as the action the ego vehicle performs in the next frame.
The downstream decoder \(q_{\psi}\) consists of a LSTM that takes the spatio-temporal embedding and final LSTM-attn hidden and cell states to decode a predicted embedding for the next time step.
This predicted embedding then goes through a final Linear layer to output an ego action class.
The supervised loss \(\mathcal{L}_{sup}\) is a \emph{Cross-Entropy} loss commonly used for multi-class classification.
Class weighting is applied when sampling batches during training to account for class imbalance (see \cref{fig:class_count}).
We will assess our methodology in terms of F1 score.

\paragraph{Similarity Retrieval}
We evaluate our embeddings, e.g. for information retrieval applications~\cite{imageRetrieval,sener2017active}, projecting them from size 20 to 2 using a Uniform Manifold Approximation and Projection (UMAP)~\cite{umap}.
We perform a qualitative analysis showing its clustering capabilities.

\paragraph{Explainability Analysis} \label{explain-method}
We apply both spatial and temporal attention to create explainability maps.
First, inspired by \cite{malawade2022roadscene2vec,spatio-temporal-attention}, we extract the node attention scores from the SAGPool layer and produce a heat map over each set of input nodes in the scene graph sequence.
Furthermore, the temporal attention scores from the LSTM-attn layer highlight which scene graph over the n-frame sequence contributed the most to the overall spatio-temporal embedding.
We analyse these maps qualitatively as the spatio-temporal attention scores should provide qualitative explanations for a user to understand how the model came to its final decision.

\section{Experimental Setup}

We investigate two methods: \textit{Pretraining and Fine-tuning} (PF) and \textit{Unsupervised Representation Learning} (URL).
Both pretrain the encoder \(f_{\theta}\) through SSL and then compose it with a downstream decoder $q_{\psi}$ under the supervision of a downstream task; PF finetunes the \(f_{\theta}\), while URL freezes its parameters while training $q_\psi$.
We compare our SSL approach with two fully-supervised models, which share the same architectural details as PF and URL: \textit{Baseline (No Aug.)} and \textit{Baseline (Aug.)}.


\paragraph{Dataset}
To train and evaluate our approach, we used the ROad event Awareness Dataset (ROAD) \cite{road-dataset}, an extension of the Oxford RobotCar Dataset~\cite{RobotCarDatasetIJRR} designed to test an AV's ability to detect road events, defined as triplets composed of an active agent, one or more actions and the corresponding scene location.
\Cref{fig:class_count} shows the available ego-vehicle actions and their strong imbalance.

\paragraph{Data Preprocessing}
Each frame is annotated with road events and their bounding box coordinates per frame.
We empirically chose a value of \(n=5\) to capture enough temporal information for each sequence.
We downsampled the video sequences 5-fold from 12 frames per second (fps) to allow the scene graphs to vary more across each 5-frame sequence, yielding about 4.5k samples.

However, one key attribute we argue is crucial to scene understanding hasn't been annotated: agents' distance from the ego vehicle.
%
Thus, we include in the graphs \textit{proximity}, calculated as the box-wise median Relative Inverse Depth (RID) resulting from a pretrained Midas (DPT Large)~\cite{midas} network, thresholded to a proximity label according to a rough estimate of its absolute depth in metres (see \cref{table: depth labels}).

\begin{table*}[htp]
    \renewcommand{\arraystretch}{1.3}
    \centering
    \begin{tabular}{l|c|c|c|c|c|c|c||c}
         &  Stop & Mov & TurRht & TurLft & MovRht & MovLft & Ovtak & Weighted Avg.  \\
         \hline
         Baseline (No Aug.) & 0.571 & 0.668 & 0.036 & 0.03 & 0.012 & 0.011 & 0.038 & 0.524 \\
         Baseline (Aug.)    & 0.771 & 0.459 & 0.257 & 0.255 & 0.137 & 0.065 & 0.306 & 0.518 \\
         URL                & 0.860 & 0.734 & 0.255 & 0.354 & 0.045 & 0.048 & 0.152 & 0.728 \\
         PF                 & 0.919 & 0.894 & 0.589 & 0.485 & 0.520 & 0.605 & 0.708 & 0.874 \\
    \end{tabular}
    \caption{Per-class and weighted F1 scores for the action prediction task.\label{tab:f1-scores}}
\end{table*}

\paragraph{Training and Network Specifications}
For pretraining, the scene graph sequences were split into a train and validation set (80/20 ratio).
As it is shown that large batch sizes greatly benefit GCL~\cite{graph-cl}, a batch size of 256 was chosen to pretrain the encoder for 50 epochs.
The Adam optimiser~\cite{adam} was used with tuned learning rate and weight decay.
Then, the encoder's weights were either frozen (for URL) or not (PF), and the downstream classifier \(q_{\psi}\) was trained, due to the limited number of scene graphs, on the same train/validation split using labels.
The models were trained for 30 epochs with a batch size of 64 with an Adam optimiser with a learning rate of \num{0.02} and weight decay rate of \num{1e-5}.
The learning rate was tuned using PyTorch Lightning's \verb|lr_find| method.

The baselines were trained for 50 epochs with a batch size of 64 and evaluated as the contrastive model.
The same technique of computing class weights was used to account for class imbalance and, in addition, \textit{Baseline (Aug.)} includes the augmentations used for SSL as in \cref{sec:contr_learning}.
The Adam optimiser was used for backpropagation with a learning rate of \num{1e-3} (tuned using PyTorch Lightning's \verb|lr_find| method) and weight decay rate of \num{5e-4}.

\paragraph{Implementation Details} \label{implement}
PyTorch Lightning and PyTorch Geometric (PyG) were the main frameworks used to build, train and validate the GNN models.
Finally, Weights \& Biases (W\&B) was used for logging metrics from experiments and performing hyperparameter sweeps.
All experiments were conducted on a server with two NVIDIA GeForce RTX 2080 Ti graphics cards.

\paragraph{Hyperparameter Tuning} \label{hyperparam}

We avoided a large hyperparameter search space by manually setting the hidden and output dimensions of the model without tuning to 64 and 20, respectively.
Each spatio-temporal embedding was projected to size 32 in the contrastive embedding space during training, and the InfoNCE \verb|tau| parameter was set to 0.5 to align with SimCLR \cite{simclr}.
We tuned \verb|num_layers|, \verb|pool_ratio|, \verb|drop_ratio|, learning rate and weight-decay rate for pretraining using W\&B sweeps, with \verb|bayes| search strategy, validation loss as metric to optimise and early termination.
A total of 13 runs were performed in the sweep.
The best-performing hyperparameters were taken to pretrain the encoder.



%

\section{Results} \label{sec:results}

\begin{figure*}[h]
    \centering
    \begin{subfigure}{0.33\textwidth}
        \includegraphics[trim=10 10 60 10,clip,width=\textwidth]{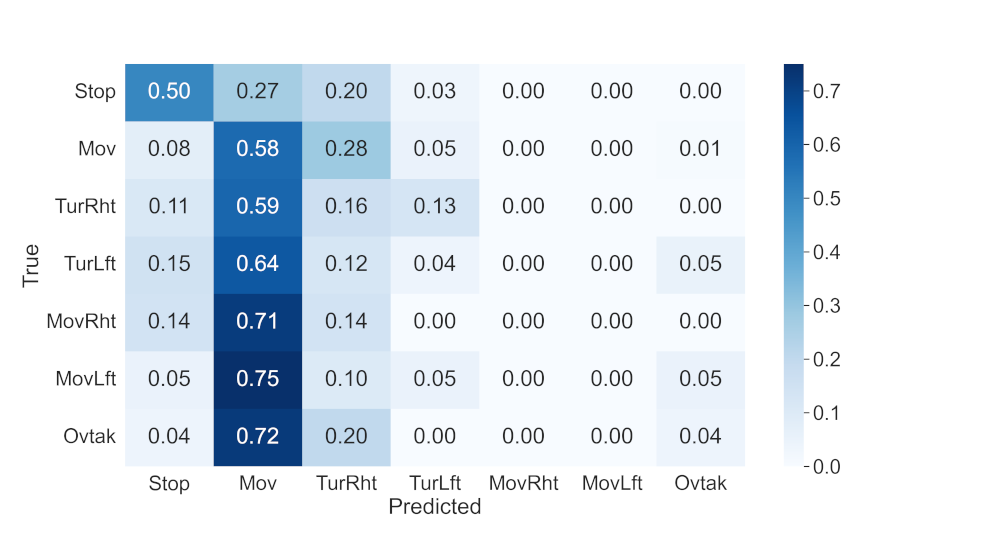}
        \caption{\textit{No Aug.}\label{fig:baseline_cm}}
    \end{subfigure}
    \hspace{20pt}
    \begin{subfigure}{0.33\textwidth}
        \includegraphics[trim=10 10 60 10,clip,width=\textwidth]{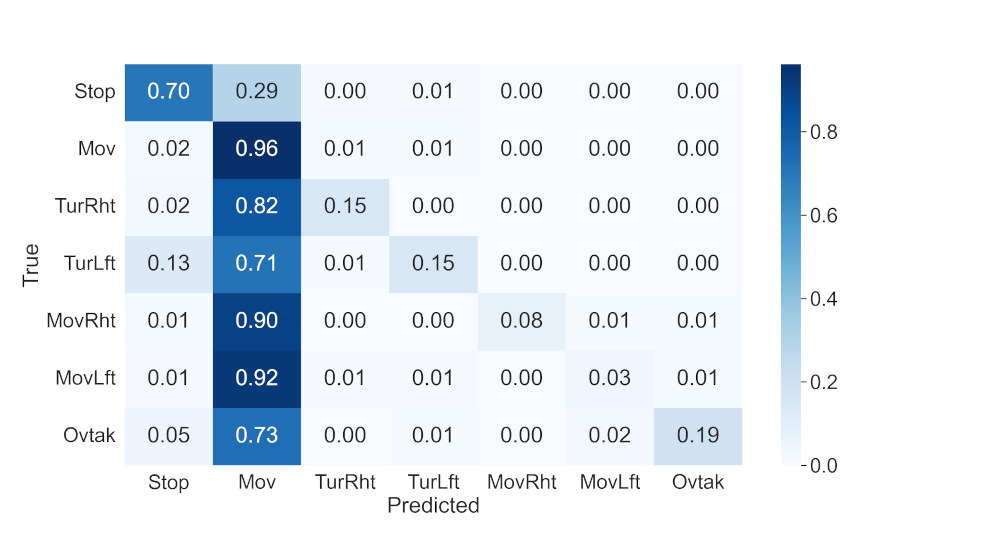}
        \caption{\textit{Aug.}\label{fig:baseline_new_cm}}
    \end{subfigure}
    
    \begin{subfigure}{0.33\textwidth}
        \includegraphics[trim=10 10 60 10, clip, width=\textwidth]{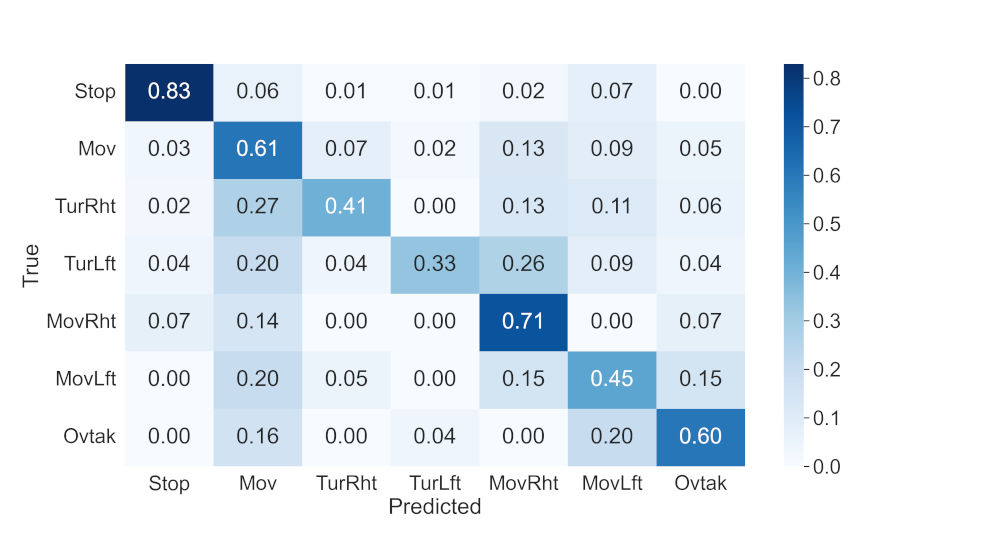}
        \caption{URL\label{fig:url_cm}}
    \end{subfigure}
    \hspace{20pt}
    \begin{subfigure}{0.33\textwidth}
        \includegraphics[trim=10 10 60 10, clip, width=\textwidth]{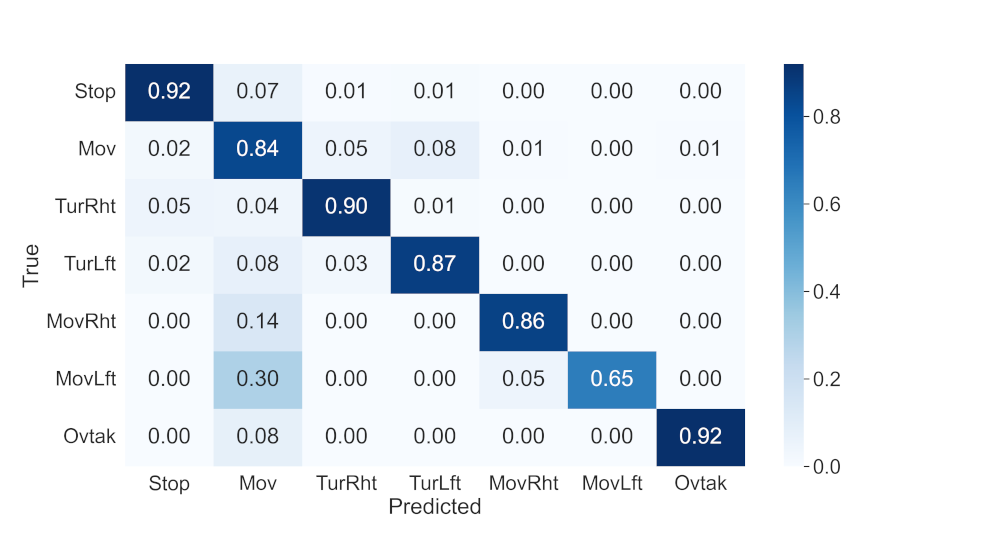}
        \caption{PF\label{fig:pf_cm}}
    \end{subfigure}

	\caption{Confusion matrices (normalised, true on row and predicted on column).\label{fig:cm}}
\end{figure*}

\paragraph{Downstream Task} \label{downstream-eval}
The performance results are presented in~\cref{tab:f1-scores} using per-class F1-scores and weighted F1-scores over the number of samples in each class.
The PF and URL contrastive models are vastly superior to the supervised baselines, especially \textit{No Aug.} which was only capable of capturing scenes that involve \emph{Move} or \emph{Stop}.
\Cref{fig:cm} shows that the supervised baselines are poor at capturing subtle differences in the scene graphs as the baseline's normalised confusion matrix only differentiates between \emph{Move} and \emph{Stop} actions, classifying most of the other actions as \emph{Move}.


It is clear from the F1-scores and normalised confusion matrices (\cref{fig:url_cm,fig:pf_cm}) that the model trained under the PF scheme outperforms the other trained under URL across all action classes.
While the two models were competitive at classifying \emph{Move} and \emph{Stop} actions, the PF model is much better at capturing scenes involving complex actions, despite the highly imbalanced classes in the dataset.

\begin{figure*}[h]
    \centering

    \begin{subfigure}{0.27\textwidth}
        \includegraphics[width=\textwidth]{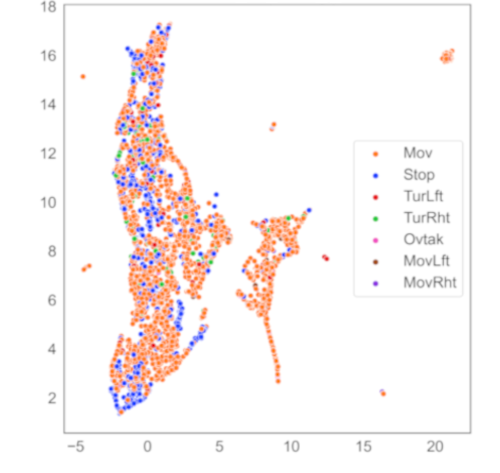}
        \caption{\textit{No Aug.} on train set\label{fig:umap_no_aug_train}}
    \end{subfigure}
    \hspace{10pt}
    \begin{subfigure}{0.27\textwidth}
        \includegraphics[width=\textwidth]{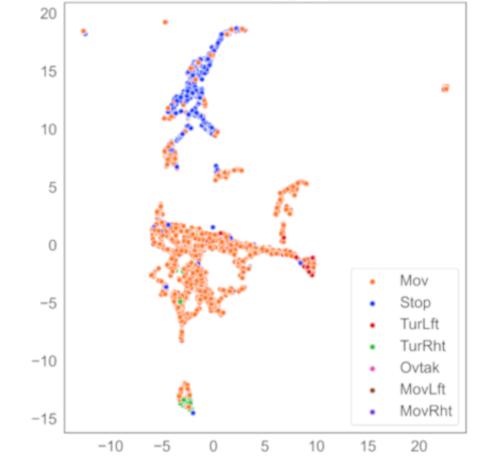}
        \caption{\textit{Aug.} on train set\label{fig:umap_aug_train}}
    \end{subfigure}
    \hspace{10pt}
    \begin{subfigure}{0.27\textwidth}
        \includegraphics[width=\textwidth]{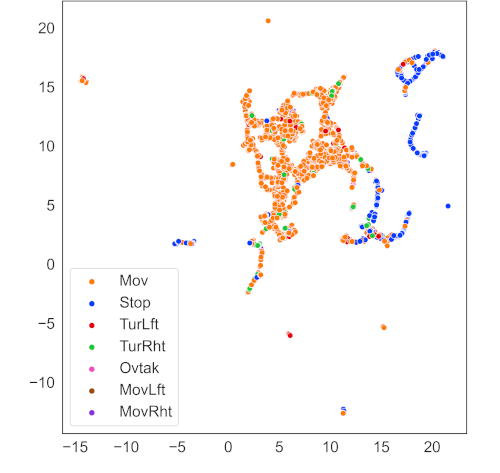}
        \caption{URL on train set\label{fig:umap_url}}
    \end{subfigure}
    
    \vspace{5pt}
    
    \begin{subfigure}{0.25\textwidth}
        \includegraphics[width=\textwidth]{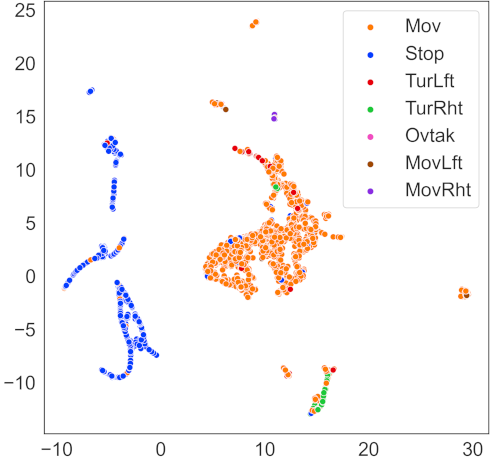}
        \caption{PF on train set\label{fig:umap_pf_train}}
    \end{subfigure}
    \hspace{20pt}
    \begin{subfigure}{0.25\textwidth}
        \includegraphics[width=\textwidth]{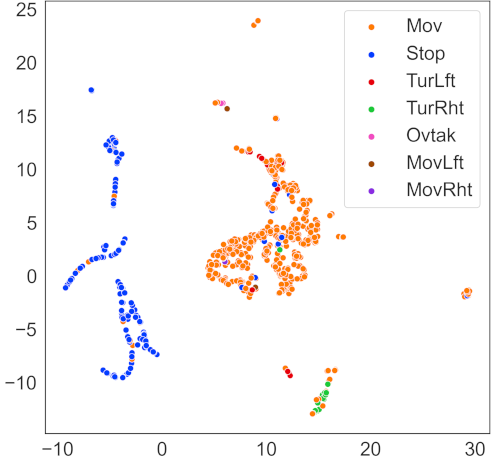}
        \caption{PF on validation set\label{fig:umap_pf_val}}
    \end{subfigure}
    \hspace{20pt}
    \begin{subfigure}{0.25\textwidth}
        \includegraphics[width=\textwidth]{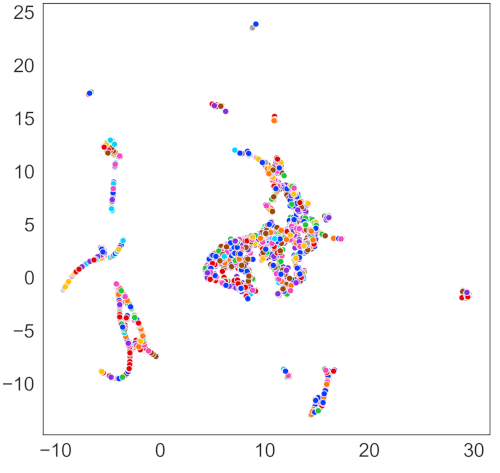}
        \caption{PF on validation set\label{fig:umap_pf_color}}
    \end{subfigure}
\caption{UMAP plots of embeddings.\label{fig:umap_overall}}
\end{figure*}

\paragraph{Similar Scene Retrieval}
\Cref{fig:umap_url,fig:umap_pf_train} assesses that URL and PF tend to cluster similar scenes based on future action labels.
The effect of imbalanced classes is clear in the baselines and URL, where embeddings for \emph{Turn Left} and \emph{Turn Right} are scattered within other main clusters.
PF does cluster \emph{Turn Right} better, as seen in the small cluster of green points at the bottom.
Interesting is the small clusters made out of \emph{Move} points outside the main cluster; a hypothesis is that these points represent edge cases that the encoder could not fully cluster.
The same behaviour can also be spotted in the validation set in~\cref{fig:umap_pf_val}.
\Cref{fig:umap_pf_color} assesses whether the embeddings are clustered solely by the future action or follow the driving video each scene graph belongs to as the 18 videos (colours in the figure) in ROAD differ in locations and times.
The embedding does not show signs of exploitation of such signals.

\begin{figure*}[h!]
    \begin{subfigure}{\textwidth}
    \centering
        \includegraphics[trim=0 112 0 0, clip, width=0.80\textwidth]{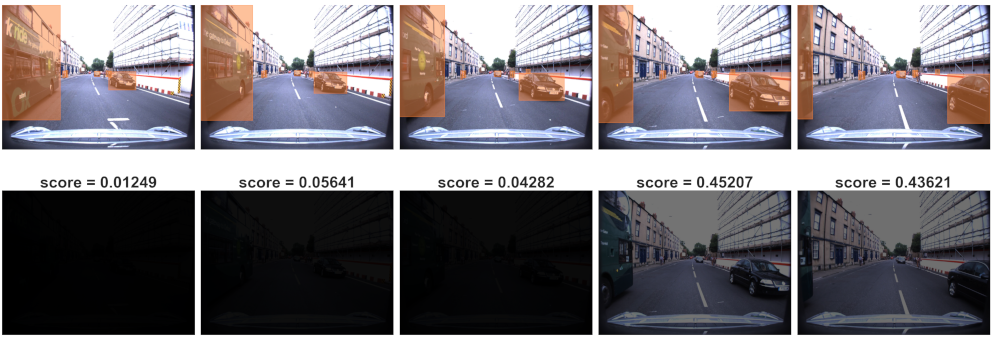}
    \caption{\label{fig:temporal_attn_viz}}
    \end{subfigure}

    \begin{subfigure}{\textwidth}
    \centering
        \includegraphics[trim=0 110 0 0, clip, width=0.80\textwidth]{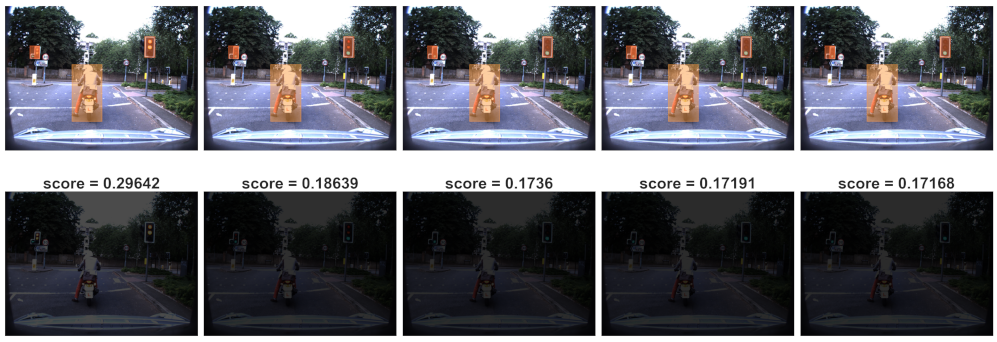}
    \caption{\label{fig:temporal_attn_mispredict}}
    \end{subfigure}
\caption{Spatial and temporal attention scores from a correctly predicted \emph{Move Left} scene (a) and a wrongly predicted \emph{Stop} scene (ground-truth \emph{Move}) (b).}
\end{figure*}

\paragraph{Explainability} \label{explain-results}
\Cref{fig:temporal_attn_viz} visualises the PF spatial and temporal attention scores over a five-frame sequence for a correctly-classified \emph{Move Left} action after overtaking a parked bus.
The model attended more to the last two frames in the sequence when predicting that the ego vehicle will steer left back to its original lane, as it would make sense as the overtaking manoeuvre nearly completes.

Explainability can also be valuable for debugging wrong prediction cases, speeding up the process of understanding where the model could have failed and ultimately improving human trust in the system.
\Cref{fig:temporal_attn_mispredict} shows a wrongly predicted scene where the ego vehicle at a junction sees the traffic light turn green.
The temporal attention scores show that the model mostly attended to the first frame where the traffic light was still amber, possibly justifying the model's behaviour.

\section{Conclusion}
This project demonstrated that spatio-temporal scene graph representations of driving scenes can be effective for AV applications, such as future action prediction, similar scene retrieval and explainability. Furthermore, the results show that scene graphs trained in a self-supervised contrastive manner outperform supervised learning methods when evaluated on a downstream task.
While this project does contain some positive results regarding the use of contrastive learning, there are undoubtedly several key limitations that exist.

\paragraph{Imbalanced Data}
The effect of an imbalanced dataset can be seen in the weighted F1-scores, where the model had a harder time correctly predicting actions such as \emph{Turn Left} or \emph{Overtake} although we applied a normalising sampling technique.
Taking inspiration from~\cite{malawade2022roadscene2vec}, a solution would be integrating synthetic samples into the training set to balance the classes.
Furthermore, due to the scalability of the scene graph extraction method, other real-world driving datasets could be integrated, such as the Honda Driving Dataset (HDD)~\cite{hdd}.

\paragraph{Encoder Design}
While using MR-GCN, SAGPool and LSTM-attn layers yielded positive results, it would be interesting to conduct an extensive ablation study on different layer types and network parameters that were not tuned for in the experiments.
Further use of attention mechanisms over edges through Relational Graph Attention Convolutions (RGATConv)~\cite{rgat} might yield better results in producing spatio-temporal embeddings for scene understanding.
Finally, it would be interesting to see if a probabilistic framework using the same encoder components benefits the task of future action prediction, as the future is inherently stochastic, as in~\cite{lecun2019}.

\paragraph{Explanations}
While we showed qualitative results for explainability, there is a lack of quantitative metrics for assessing this capability, as considerable human interpretation is required to judge the explanations when verifying or debugging model predictions.
Taking inspiration from \cite{explain-gcn}, two quantitative metrics for GCN explanations can be used, \emph{fidelity} and \emph{contrastivity}, each designed to capture specific properties.
Fidelity aims to capture the intuition that the occlusion of salient nodes identified through explanations should decrease classification accuracy.
Contrastivity, however, captures the intuition that class-specific features highlighted by an explanation should differ between classes.
\section*{Acknowledgements}

This work was supported by the EPSRC project RAILS (grant reference: EP/W011344/1), the EPSRC Programme Grant ``From Sensing to Collaboration'' (EP/V000748/1) and the ORI research project RobotCycle.

\bibliographystyle{IEEEtran}
\bibliography{mybib}

\begin{thebibliography}{10}
\providecommand{\url}[1]{#1}
\csname url@samestyle\endcsname
\providecommand{\newblock}{\relax}
\providecommand{\bibinfo}[2]{#2}
\providecommand{\BIBentrySTDinterwordspacing}{\spaceskip=0pt\relax}
\providecommand{\BIBentryALTinterwordstretchfactor}{4}
\providecommand{\BIBentryALTinterwordspacing}{\spaceskip=\fontdimen2\font plus
\BIBentryALTinterwordstretchfactor\fontdimen3\font minus
  \fontdimen4\font\relax}
\providecommand{\BIBforeignlanguage}[2]{{%
\expandafter\ifx\csname l@#1\endcsname\relax
\typeout{** WARNING: IEEEtran.bst: No hyphenation pattern has been}%
\typeout{** loaded for the language `#1'. Using the pattern for}%
\typeout{** the default language instead.}%
\else
\language=\csname l@#1\endcsname
\fi
#2}}
\providecommand{\BIBdecl}{\relax}
\BIBdecl

\bibitem{Omeiza_2021}
\BIBentryALTinterwordspacing
D.~Omeiza, H.~Webb, M.~Jirotka, and L.~Kunze, ``Explanations in autonomous
  driving: A survey,'' \emph{{IEEE} Transactions on Intelligent Transportation
  Systems}, pp. 1--21, 2021. [Online]. Available:
  \url{https://doi.org/10.1109%2Ftits.2021.3122865}
\BIBentrySTDinterwordspacing

\bibitem{relational_bias}
\BIBentryALTinterwordspacing
P.~W. Battaglia, J.~B. Hamrick, V.~Bapst, A.~Sanchez-Gonzalez, V.~Zambaldi,
  M.~Malinowski, A.~Tacchetti, D.~Raposo, A.~Santoro, R.~Faulkner, C.~Gulcehre,
  F.~Song, A.~Ballard, J.~Gilmer, G.~Dahl, A.~Vaswani, K.~Allen, C.~Nash,
  V.~Langston, C.~Dyer, N.~Heess, D.~Wierstra, P.~Kohli, M.~Botvinick,
  O.~Vinyals, Y.~Li, and R.~Pascanu, ``Relational inductive biases, deep
  learning, and graph networks,'' 2018. [Online]. Available:
  \url{https://arxiv.org/abs/1806.01261}
\BIBentrySTDinterwordspacing

\bibitem{vehicle_cls}
\BIBentryALTinterwordspacing
S.~Mylavarapu, M.~Sandhu, P.~Vijayan, K.~M. Krishna, B.~Ravindran, and
  A.~Namboodiri, ``Towards accurate vehicle behaviour classification with
  multi-relational graph convolutional networks,'' 2020. [Online]. Available:
  \url{https://arxiv.org/abs/2002.00786}
\BIBentrySTDinterwordspacing

\bibitem{collision_pred}
\BIBentryALTinterwordspacing
A.~V. Malawade, S.-Y. Yu, B.~Hsu, D.~Muthirayan, P.~P. Khargonekar, and
  M.~A.~A. Faruque, ``Spatio-temporal scene-graph embedding for autonomous
  vehicle collision prediction,'' 2021. [Online]. Available:
  \url{https://arxiv.org/abs/2111.06123}
\BIBentrySTDinterwordspacing

\bibitem{risk_assess}
\BIBentryALTinterwordspacing
S.-Y. Yu, A.~V. Malawade, D.~Muthirayan, P.~P. Khargonekar, and M.~A.~A.
  Faruque, ``Scene-graph augmented data-driven risk assessment of autonomous
  vehicle decisions,'' 2020. [Online]. Available:
  \url{https://arxiv.org/abs/2009.06435}
\BIBentrySTDinterwordspacing

\bibitem{graph-ssl}
\BIBentryALTinterwordspacing
Y.~Liu, S.~Pan, M.~Jin, C.~Zhou, F.~Xia, and P.~S. Yu, ``Graph self-supervised
  learning: {A} survey,'' \emph{CoRR}, vol. abs/2103.00111, 2021. [Online].
  Available: \url{https://arxiv.org/abs/2103.00111}
\BIBentrySTDinterwordspacing

\bibitem{3d-aware-scenegraph}
\BIBentryALTinterwordspacing
C.~Li, Y.~Meng, S.~H. Chan, and Y.~Chen, ``Learning 3d-aware egocentric
  spatial-temporal interaction via graph convolutional networks,'' \emph{CoRR},
  vol. abs/1909.09272, 2019. [Online]. Available:
  \url{http://arxiv.org/abs/1909.09272}
\BIBentrySTDinterwordspacing

\bibitem{dynamic_scene_understanding}
\BIBentryALTinterwordspacing
S.~Mylavarapu, M.~Sandhu, P.~Vijayan, K.~M. Krishna, B.~Ravindran, and
  A.~Namboodiri, ``Understanding dynamic scenes using graph convolution
  networks,'' \emph{CoRR}, vol. abs/2005.04437, 2020. [Online]. Available:
  \url{https://arxiv.org/abs/2005.04437}
\BIBentrySTDinterwordspacing

\bibitem{malawade2022roadscene2vec}
\BIBentryALTinterwordspacing
A.~V. Malawade, S.-Y. Yu, B.~Hsu, H.~Kaeley, A.~Karra, and M.~A. {Al Faruque},
  ``roadscene2vec: A tool for extracting and embedding road scene-graphs,''
  \emph{Knowledge-Based Systems}, vol. 242, p. 108245, 2022. [Online].
  Available:
  \url{https://www.sciencedirect.com/science/article/pii/S0950705122000739}
\BIBentrySTDinterwordspacing

\bibitem{gnn-survey}
\BIBentryALTinterwordspacing
Z.~Wu, S.~Pan, F.~Chen, G.~Long, C.~Zhang, and P.~S. Yu, ``A comprehensive
  survey on graph neural networks,'' \emph{CoRR}, vol. abs/1901.00596, 2019.
  [Online]. Available: \url{http://arxiv.org/abs/1901.00596}
\BIBentrySTDinterwordspacing

\bibitem{group-activity-recog}
\BIBentryALTinterwordspacing
J.~Wu, L.~Wang, L.~Wang, J.~Guo, and G.~Wu, ``Learning actor relation graphs
  for group activity recognition,'' \emph{CoRR}, vol. abs/1904.10117, 2019.
  [Online]. Available: \url{http://arxiv.org/abs/1904.10117}
\BIBentrySTDinterwordspacing

\bibitem{scene-graph-gen}
\BIBentryALTinterwordspacing
Y.~Cong, W.~Liao, H.~Ackermann, M.~Y. Yang, and B.~Rosenhahn,
  ``Spatial-temporal transformer for dynamic scene graph generation,''
  \emph{CoRR}, vol. abs/2107.12309, 2021. [Online]. Available:
  \url{https://arxiv.org/abs/2107.12309}
\BIBentrySTDinterwordspacing

\bibitem{action-recog}
\BIBentryALTinterwordspacing
W.~Xie, J.~K. Chen, and A.~Z. Luo, ``Towards compositional action recognition
  with spatio-temporal graph neural network.'' [Online]. Available:
  \url{https://russellxie7.me/docs/graph.pdf}
\BIBentrySTDinterwordspacing

\bibitem{graph-cl}
\BIBentryALTinterwordspacing
Y.~You, T.~Chen, Y.~Sui, T.~Chen, Z.~Wang, and Y.~Shen, ``Graph contrastive
  learning with augmentations,'' \emph{CoRR}, vol. abs/2010.13902, 2020.
  [Online]. Available: \url{https://arxiv.org/abs/2010.13902}
\BIBentrySTDinterwordspacing

\bibitem{grace}
\BIBentryALTinterwordspacing
Y.~Zhu, Y.~Xu, F.~Yu, Q.~Liu, S.~Wu, and L.~Wang, ``Deep graph contrastive
  representation learning,'' \emph{CoRR}, vol. abs/2006.04131, 2020. [Online].
  Available: \url{https://arxiv.org/abs/2006.04131}
\BIBentrySTDinterwordspacing

\bibitem{infograph}
\BIBentryALTinterwordspacing
F.~Sun, J.~Hoffmann, and J.~Tang, ``Infograph: Unsupervised and semi-supervised
  graph-level representation learning via mutual information maximization,''
  \emph{CoRR}, vol. abs/1908.01000, 2019. [Online]. Available:
  \url{http://arxiv.org/abs/1908.01000}
\BIBentrySTDinterwordspacing

\bibitem{simclr}
\BIBentryALTinterwordspacing
T.~Chen, S.~Kornblith, M.~Norouzi, and G.~E. Hinton, ``A simple framework for
  contrastive learning of visual representations,'' \emph{CoRR}, vol.
  abs/2002.05709, 2020. [Online]. Available:
  \url{https://arxiv.org/abs/2002.05709}
\BIBentrySTDinterwordspacing

\bibitem{explain-gcn}
P.~E. Pope, S.~Kolouri, M.~Rostami, C.~E. Martin, and H.~Hoffmann,
  ``Explainability methods for graph convolutional neural networks,'' in
  \emph{Proceedings of the IEEE/CVF Conference on Computer Vision and Pattern
  Recognition (CVPR)}, June 2019.

\bibitem{gin}
\BIBentryALTinterwordspacing
K.~Xu, W.~Hu, J.~Leskovec, and S.~Jegelka, ``How powerful are graph neural
  networks?'' \emph{CoRR}, vol. abs/1810.00826, 2018. [Online]. Available:
  \url{http://arxiv.org/abs/1810.00826}
\BIBentrySTDinterwordspacing

\bibitem{sagpool}
\BIBentryALTinterwordspacing
J.~Lee, I.~Lee, and J.~Kang, ``Self-attention graph pooling,'' \emph{CoRR},
  vol. abs/1904.08082, 2019. [Online]. Available:
  \url{http://arxiv.org/abs/1904.08082}
\BIBentrySTDinterwordspacing

\bibitem{seq2seq}
\BIBentryALTinterwordspacing
D.~Bahdanau, K.~Cho, and Y.~Bengio, ``Neural machine translation by jointly
  learning to align and translate,'' 2014. [Online]. Available:
  \url{https://arxiv.org/abs/1409.0473}
\BIBentrySTDinterwordspacing

\bibitem{jse}
\BIBentryALTinterwordspacing
P.~Veličković, W.~Fedus, W.~L. Hamilton, P.~Liò, Y.~Bengio, and R.~D. Hjelm,
  ``Deep graph infomax,'' 2018. [Online]. Available:
  \url{https://arxiv.org/abs/1809.10341}
\BIBentrySTDinterwordspacing

\bibitem{imageRetrieval}
\BIBentryALTinterwordspacing
B.~Schroeder and S.~Tripathi, ``Structured query-based image retrieval using
  scene graphs,'' \emph{CoRR}, vol. abs/2005.06653, 2020. [Online]. Available:
  \url{https://arxiv.org/abs/2005.06653}
\BIBentrySTDinterwordspacing

\bibitem{sener2017active}
O.~Sener and S.~Savarese, ``Active learning for convolutional neural networks:
  A core-set approach,'' \emph{arXiv preprint arXiv:1708.00489}, 2017.

\bibitem{umap}
\BIBentryALTinterwordspacing
L.~McInnes, J.~Healy, and J.~Melville, ``Umap: Uniform manifold approximation
  and projection for dimension reduction,'' 2018. [Online]. Available:
  \url{https://arxiv.org/abs/1802.03426}
\BIBentrySTDinterwordspacing

\bibitem{spatio-temporal-attention}
\BIBentryALTinterwordspacing
L.~Meng, B.~Zhao, B.~Chang, G.~Huang, F.~Tung, and L.~Sigal, ``Where and when
  to look? spatio-temporal attention for action recognition in videos,''
  \emph{CoRR}, vol. abs/1810.04511, 2018. [Online]. Available:
  \url{http://arxiv.org/abs/1810.04511}
\BIBentrySTDinterwordspacing

\bibitem{road-dataset}
\BIBentryALTinterwordspacing
G.~Singh, S.~Akrigg, M.~D. Maio, V.~Fontana, R.~J. Alitappeh, S.~Saha, K.~J.
  Saravi, F.~Yousefi, J.~Culley, T.~Nicholson, J.~Omokeowa, S.~Khan,
  S.~Grazioso, A.~Bradley, G.~D. Gironimo, and F.~Cuzzolin, ``{ROAD:} the road
  event awareness dataset for autonomous driving,'' \emph{CoRR}, vol.
  abs/2102.11585, 2021. [Online]. Available:
  \url{https://arxiv.org/abs/2102.11585}
\BIBentrySTDinterwordspacing

\bibitem{RobotCarDatasetIJRR}
\BIBentryALTinterwordspacing
W.~Maddern, G.~Pascoe, C.~Linegar, and P.~Newman, ``{1 Year, 1000km: The Oxford
  RobotCar Dataset},'' \emph{The International Journal of Robotics Research
  (IJRR)}, vol.~36, no.~1, pp. 3--15, 2017. [Online]. Available:
  \url{http://dx.doi.org/10.1177/0278364916679498}
\BIBentrySTDinterwordspacing

\bibitem{midas}
\BIBentryALTinterwordspacing
K.~Lasinger, R.~Ranftl, K.~Schindler, and V.~Koltun, ``Towards robust monocular
  depth estimation: Mixing datasets for zero-shot cross-dataset transfer,''
  \emph{CoRR}, vol. abs/1907.01341, 2019. [Online]. Available:
  \url{http://arxiv.org/abs/1907.01341}
\BIBentrySTDinterwordspacing

\bibitem{adam}
\BIBentryALTinterwordspacing
D.~P. Kingma and J.~Ba, ``Adam: A method for stochastic optimization,'' 2014.
  [Online]. Available: \url{https://arxiv.org/abs/1412.6980}
\BIBentrySTDinterwordspacing

\bibitem{hdd}
\BIBentryALTinterwordspacing
V.~Ramanishka, Y.~Chen, T.~Misu, and K.~Saenko, ``Toward driving scene
  understanding: {A} dataset for learning driver behavior and causal
  reasoning,'' \emph{CoRR}, vol. abs/1811.02307, 2018. [Online]. Available:
  \url{http://arxiv.org/abs/1811.02307}
\BIBentrySTDinterwordspacing

\bibitem{rgat}
\BIBentryALTinterwordspacing
D.~Busbridge, D.~Sherburn, P.~Cavallo, and N.~Y. Hammerla, ``Relational graph
  attention networks,'' \emph{CoRR}, vol. abs/1904.05811, 2019. [Online].
  Available: \url{http://arxiv.org/abs/1904.05811}
\BIBentrySTDinterwordspacing

\bibitem{lecun2019}
\BIBentryALTinterwordspacing
M.~Henaff, A.~Canziani, and Y.~LeCun, ``Model-predictive policy learning with
  uncertainty regularization for driving in dense traffic,'' \emph{CoRR}, vol.
  abs/1901.02705, 2019. [Online]. Available:
  \url{http://arxiv.org/abs/1901.02705}
\BIBentrySTDinterwordspacing

\end{thebibliography}

\end{document}